\documentclass[10pt,twocolumn,letterpaper]{article}

\usepackage{iccv}
\usepackage{times}
\usepackage{epsfig}
\usepackage{graphicx}
\usepackage{amsmath}
\usepackage{amssymb}
\usepackage{bbm}
\usepackage[accsupp]{axessibility}  % Improves PDF readability for those with disabilities.

\newcommand{\relmix}{RelMix}

\newcommand{\vx}{\textbf{x}}
\newcommand{\vy}{\textbf{y}}

\usepackage{subcaption}
\usepackage{amsthm}
\usepackage{booktabs}
\usepackage{mathtools}

\usepackage[pagebackref=true,breaklinks=true,letterpaper=true,colorlinks,bookmarks=false]{hyperref}

\iccvfinalcopy % *** Uncomment this line for the final submission

\def\iccvPaperID{10473} % *** Enter the ICCV Paper ID here

\ificcvfinal\fi

\begin{document}

%%%%%%%%% TITLE
\title{Exploring Long Tail Visual Relationship Recognition with Large Vocabulary}% \\ with Hubless Regularized RelMix}

% \author{Sherif Abdelkarim\thanks{Equal contribution}  \thanks{Work done while working at King Abdullah University of Science and Technology}\\
% KAUST, University of California Irvine\\
% {\tt\small abdelkas@uci.edu}
% \and
% Aniket Agarwal\footnotemark[1] \footnotemark[2]\\
% KAUST, Indian Institute of Technology\\
% {\tt\small aagarwal@ma.iitr.ac.in}
% \and
% Panos Achlioptas\\
% Stanford University\\
% {\tt\small panos@cs.stanford.edu}
% \and
% Jun Chen\\
% %King Abdullah University of Science and Technology\\
% KAUST\\
% {\tt\small jun.chen@kaust.edu.sa}
% \and
% Jiaji Haung\\
% Baidu SVAIL\\
% {\tt\small  huangjiaji@baidu.com}
% \and
% Boyang Lee\\
% Nanyang Technological University\\
% {\tt\small lily.liboyang@ntu.edu.sg}
% \and
% Kenneth Church\\
% Baidu SVAIL\\
% {\tt\small kennethchurch@baidu.com}
% \and
% Mohamed Elhosseiny\\
% KAUST, Stanford University\\
% {\tt\small mohamed.elhoseiny@kaust.edu.sa}
% }

\author{Sherif Abdelkarim\textsuperscript{\rm 1}\thanks{Equal contribution} \thanks{Work done while working at King Abdullah University of Science and Technology (KAUST)} \thanks{Corresponding Authors} , Aniket Agarwal\textsuperscript{\rm 1,2}\footnotemark[1] \footnotemark[2], Panos Achlioptas\textsuperscript{\rm 3}, Jun Chen\textsuperscript{\rm 1} \\ Jiaji Huang\textsuperscript{\rm 4}, Boyang Li\textsuperscript{\rm 5}, Kenneth Church\textsuperscript{\rm 4}, Mohamed Elhoseiny\textsuperscript{\rm 1}\footnotemark[3] \\

\textsuperscript{\rm 1}{\small King Abdullah University of Science and Technology (KAUST)},  %If you have multiple authors and multiple affiliations
\textsuperscript{\rm 2}{\small IIT Roorkee}, 
\textsuperscript{\rm 3}{\small Stanford University}, 
\textsuperscript{\rm 4}{\small Baidu}, 
\textsuperscript{\rm 5}{\small NTU Singapore} \\

{\tt\footnotesize abdelkas@uci.edu}, ~{\tt\footnotesize aagarwal@ma.iitr.ac.in},  {\tt\footnotesize panos@cs.stanford.edu}, ~{\tt\footnotesize jun.chen@kaust.edu.sa}, \\~{\tt\footnotesize  huangjiaji@baidu.com}, ~{\tt\footnotesize lily.liboyang@ntu.edu.sg}, ~{\tt\footnotesize kennethchurch@baidu.com}, ~{\tt\footnotesize mohamed.elhoseiny@kaust.edu.sa}
}

\maketitle
\ificcvfinal\thispagestyle{empty}\fi

%%%%%%%%% ABSTRACT
\begin{abstract}

% Several approaches have been proposed in recent literature to alleviate the long-tail problem, mainly in object classification tasks. In this paper, we make the first large-scale study concerning the task of Long-Tail Visual Relationship Recognition (LTVRR). LTVRR aims at improving the learning of structured visual relationships that come from the long-tail (e.g.,  ``\textit{rabbit grazing on grass}"). In this setup,  subject, relation, and object classes individually follow a  long-tail distribution. We first introduce two large-scale long-tail visual relationship recognition benchmarks to study this task, dubbed VG8K-LT and GQA-LT. VG8K-LT and GQA-LT are built upon the widely used Visual Genome and GQA datasets. 
% We use these benchmarks to study the performance of several state-of-the-art long-tail models on the LTVRR setup. Also, we propose a visiolinguistic hubless (VilHub) loss and a Mixup augmentation technique adapted to LTVRR setup, dubbed as RelMix.  Both VilHub and RelMix can be easily integrated on top of existing models and despite being simple, our results show that they can remarkably improve the performance, especially on tail classes.
% The benchmarks and code will be made available.

Several approaches have been proposed in recent literature to alleviate the long-tail problem, mainly in object classification tasks. In this paper, we make the first large-scale study concerning the task of Long-Tail Visual Relationship Recognition (LTVRR). LTVRR aims at improving the learning of structured visual relationships that come from the long-tail (e.g., “\textit{rabbit grazing on grass}”). In this setup, the subject, relation, and object classes each follow a long-tail distribution. To begin our study and make a future benchmark for the community, we introduce two LTVRR-related benchmarks, dubbed VG8K-LT and GQA-LT, built upon the widely used Visual Genome and GQA datasets. We use these benchmarks to study the performance of several state-of-the-art long-tail models on the LTVRR setup. Lastly, we propose a visiolinguistic hubless (VilHub) loss and a Mixup augmentation technique adapted to LTVRR setup, dubbed as RelMix. Both VilHub and RelMix can be easily integrated on top of existing models and despite being simple, our results show that they can remarkably improve the performance, especially on tail classes.
Benchmarks, code, and models have been made available at: \href{https://github.com/Vision-CAIR/LTVRR}{https://github.com/Vision-CAIR/LTVRR}.
\end{abstract}

%%%%%%%%% BODY TEXT

\section{Introduction}
\label{sec_intro}
% \begin{figure}[ht]
% \centering
%  \includegraphics[width=\linewidth]{figures/relmix.png}
%  \caption{This figures shows the basic mechanism behind RelMix. On the left hand side is the original image, and on the right hand side is the augmented image, after replacing \emph{man} with \emph{cat}}
% \label{fig:teaser}
% \end{figure}

%% We also show relmix strategy is helpful in this setting below the figure
\begin{figure}[t]
\centering
 \includegraphics[width=.8\linewidth]{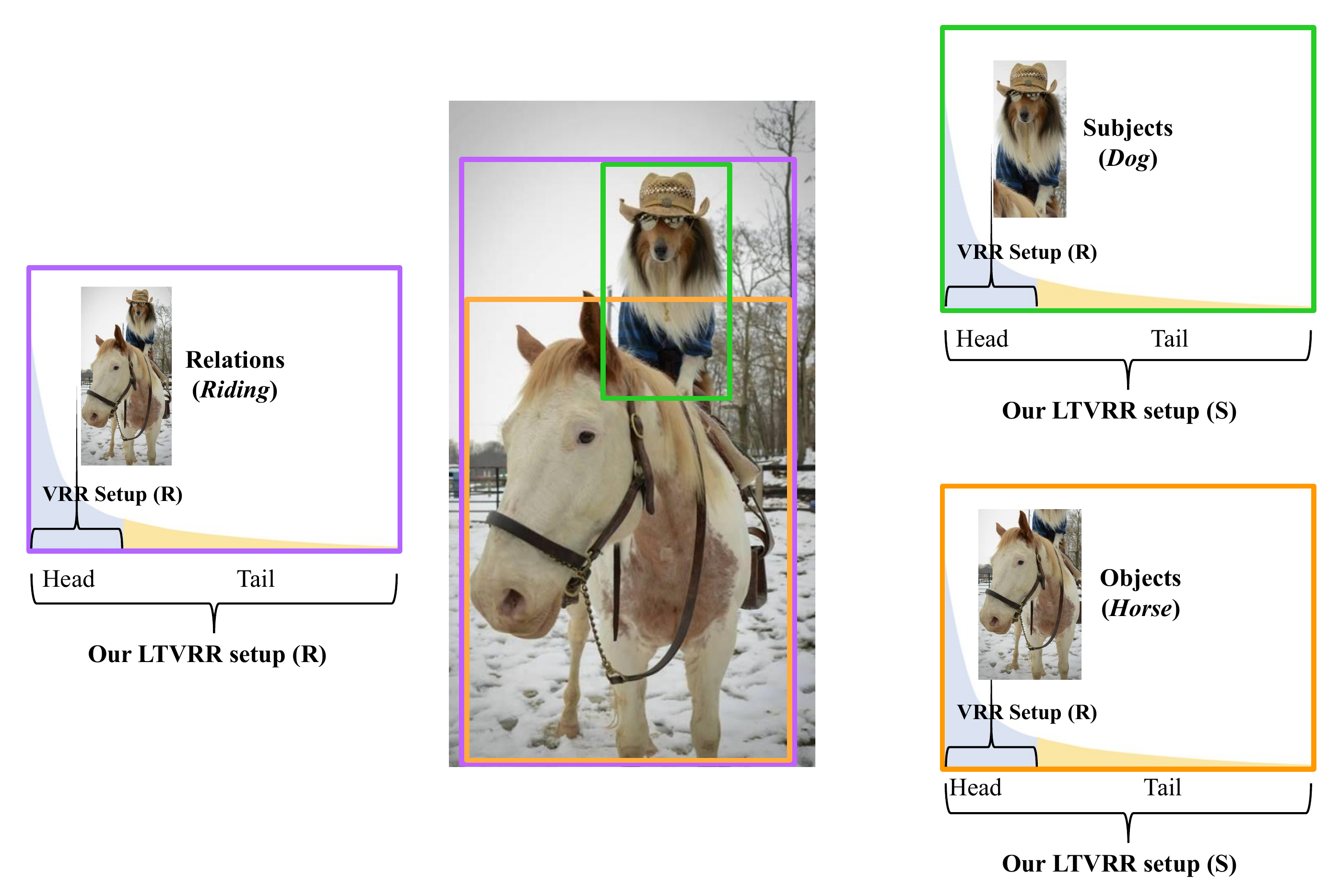}
   \caption{\textbf{L}ong-\textbf{T}ail \textbf{V}isual \textbf{R}elationship \textbf{R}ecognition (\textit{LTVRR}). In contrast to existing Visual Relationship Recognition (VRR) setups, where each of the subject (S), objects(O), and relationships (R) have abundant examples,  In \textit{LTVRR}, we focus on such rare relational events where S,R, and O also follow a long-tail distribution which we believe is more realistic and challenging}
%   , yet the long-tailness gets more severe with compositions and requires more explicit modeling to improve performance. We also introduce visiolinguistic hubless regularization and RelMix augmentation to improve \textit{LTVRR} performance among other baselines. 
\label{fig:teaser}
\end{figure}

% In LTVRR, we study the task of recognizing visual relationships in images by identifying the interacting objects and the relationship between them where the subject, relation, and object follow a long-tail distribution~\cite{zipf1949human}.
%Since AlexNet \cite{krizhevsky2012imagenet} and the ImageNet Large Scale Visual Recognition Challenge \cite{russakovsky2014imagenet}, the task of visual recognition has witnessed significant progress, often being described as surpassing human performance.  However, 

% Research on visual recognition has been largely focused on cases where training data is abundant ~\cite{deng2009imagenet,krizhevsky2012imagenet,simonyan2014very,he2016deep,huang2017densely}. However, real world distributions usually resemble the power-law (long-tail) distribution, where a small portion of the categories contain most of the data. Even though The long-tail problem has been recognized in the literature~\cite{zipf1949human} for a long time, the research is not as extensive~\cite{salakhutdinov2011learning} as it is on balanced data distributions.

Most existing works in visual recognition assume that training data are abundant, with typically a few hundred to thousands of examples per class~\cite{deng2009imagenet,krizhevsky2012imagenet,simonyan2014very,he2016deep,huang2017densely}. A more realistic setup, however, is to assume that  classes follow a long-tail distribution, where most categories have only few examples. What makes the long-tail distribution more natural is that it covers the spectrum of frequent classes, few-shot classes (classes rarely observed in the training set) and even zero-shot classes (classes that do not appear in the training set). 
Few-shot and zero-shot learning has been separately studied in~\cite{snell2017prototypical,ren2018meta,zhang2018metagan} and~\cite{xian2018feature,xian2018feature,gbu}, respectively. 

\noindent \textbf{ LTVRR}
% While the long-tail problem has been recognized in language research for a long time (e.g., ~\cite{zipf1949human}), the literature on the long-tail in vision is relatively more recent (e.g., ~\cite{salakhutdinov2011learning}).
Several approaches have been developed to advance Long-Tail Object Recognition (LTOR) \cite{lin2017focal,liu2019large,wang2017learning,9157168}.  However, most of the metrics and evaluation setups in long-tail object recognition do not apply to the Visual Relationship Recognition (\emph{VRR}) literature, which is more complex and structured. The goal of the \emph{VRR} task is to recognize the categories of two interacting objects and their relation, e.g., recognizing triplets like $<$dog, riding, horse$>$~\cite{lu2016visual,plummerPLCLC2017,zhang2017relationship}.  In contrast to most existing \emph{VRR} benchmarks,   all object categories no matter their frequency  contribute equally to evaluation metrics in \emph{LTOR}, where the average per class accuracy is the common metric. Inspired by \emph{LTOR} literature, we  extend their long-tail setup to study visual relationship recognition. In our setup, dubbed {Long-Tail Visual Relationship Recognition} (\emph{LTVRR}), subjects, objects, and relationships follow a long-tail distribution; see Fig~\ref{fig:teaser}. In this setup, this structured recognition task is more challenging as not only could the combination (S, R, O) be rare, but so can any of the interacting subjects/objects (S/O) and/or the relation (R). 
An important distinction between our and previous works is that our focus is on much more long-tailed distributions than previous methods. Most long-tail literature focuses on the range of class frequency that it is on a smaller scale than in our setup (between $5$ and $5000$ for ~\cite{liu2019large}, between $1$ and $1000$ for~\cite{lu2016visual}, and  which is around a factor of $1000$ between the most frequent and the least frequent classes). On the other hand, for our benchmarks, we use the following range of frequencies:
For GQA-LT ( $1,703$ object classes and $310$ relation classes), the most frequent object and relationship categories have $374,282$ and $1,692,068$ examples, and the least frequent have $1$ and $2$ examples, respectively. This results in factors of around $300,000+$ for objects and around $1.7$ million for relations between the most frequent and least frequent classes.
For VG8K-LT ($5,330$ objects classes and $2000$ relation classes), the most frequent object and relationship categories have $196,944$ and $618,687$ examples, and the least frequent have $14$ and $18$ examples, respectively, which leads to factors of approximately $14,000$ for objects and $34,000$ for relations; 
% This contrast is illustrated visually in Fig.~\ref{fig:teaser}.
see more details in Sec.~\ref{sec:experiments}

%noindent 
%\textbf{ Augmentation in Object Recognition.} 

We also implement several state-of-the-art models \cite{9157168, lin2017focal, Kang2020Decoupling, liu2019large} targetted on long-tail object classification in our LTVRR setup, which we believe is crucial for further work on this setup. Orthogonally, we also propose a novel augmentation technique, dubbed RelMix and a hubless regularization loss, introduced in section~\ref{sec_approach}. Inspired from \cite{pmlr-v97-verma19a}, in \relmix, we augment the training data systematically using a combination of features to improve upon the tail performance. This effectively helps in augmenting more data for tail classes, hence balancing the head and tail distribution. We also regularize the model by casting long-tail visual understanding as a hubness problem and introduce a Visio-linguistic Hubless (VilHub) loss. The approach is inspired by hubness literature in Natural Language Processing (e.g.,  \cite{huang2019hubless,lazaridou-etal-2015-hubness}) but differs in (a) they use the hubness to improve word-level translation from one language to another, at the same time we model hubness in a visio-lingual task connecting vision to language.  (b) Our approach can correct learning representation that minimizes hubness from deep vision and language neural networks in an end-to-end way in contrast to only correcting bias parameters~\cite{huang2019hubless}.

\noindent \textbf{Contributions:} \\
\noindent (1) We adapt several state of the art approaches in long-tail classification to our setup and report the performances on two proposed benchmarks GQA-LT and VG8K-LT. Due to the large vocabulary size of  objects and relationships in the \emph{LTVRR} setup, we also analyze the models based on their capacity to bring categories that are semantically similar to the ground-truth, higher in the rank of the model's predictions according to wordNet~\cite{miller1995wordnet}, and word2vec~\cite{zhang2019large}. We found this to be useful, especially when the vocabulary of predictions is large. \\  %\\  \jun{these two datasets focus on the long-tail or it exists the severe long-tail problem?}\\
\noindent (2) We propose a novel augmentation method, dubbed RelMix, for the visual relationship recognition problem. We empirically show that our augmentation method, while simple, effectively improves the performance across the whole class distribution with more focus on tail classes.\\
\noindent (3)  %\mohamed{mentioned it but emphasize less.move detailed contrast  to related work. } 
We propose to cast the long-tail visual
understanding as a hubness problem, and introduce a Visio-linguistic Hubless (VilHub) loss.% Our approach is inspired from hubness literature in Natural Language Processing (e.g.,  \cite{huang2019hubless,lazaridou-etal-2015-hubness}), but differs in (a) they use the hubness to improve word-level translation from one language to another, while we model hubness in a visio-lingual task connecting vision to language.  (b) Our approach can correct learning representation that minimizes hubness from both deep vision and language neural networks in an end-to-end fashion in contrast to bias correction losses adopted in NLP domain~\cite{huang2019hubless}. %\jun{don't really understand what is the meaning here. what problems you solved for hubless loss function on NLP task?}.
We showed that VilHub loss can be simply integrated with some existing losses like Focal Loss (FL)~\cite{lin2017focal} and Weighted Cross Entropy~\cite{lin2017focal} to improve performance as an effective regularizer.\\ % todo search for better citation for WCE

\section{Related Work}
\label{sec_related_work}
\label{sec:related}
 
% -------------------------------------------
 
\noindent \textbf{Visual Relationship Detection} %todo add references to the statement 'Visual relationship detection has been extensively studied'
Visual relationship detection (VRD) has been extensively studied in the past few years~\cite{lu2016visual,zhang2017visual,xu2017scenegraph,lu2016visual}. 
% Most of the methods used are based on a small vocabulary, e.g., 100 objects and 70 relations from VRD dataset~\cite{lu2016visual}, or a portion of Visual Genome dataset, VG200 which contains 150 objects and 50 relations of the most frequent objects and relations~\cite{zhang2017visual,xu2017scenegraph}.
Lu et.al.~\cite{lu2016visual} utilizes the object detection output of an R-CNN detector and leverages language priors for relationship prediction.
% Although language embeddings are used, they are only taken into account for ranking the relations in a ranking loss framework.
% ~\cite{Zhuang_2017_ICCV} use language representations of the subject and object as contexts to improve relation prediction with a pre-trained language representation.
%Very recently, 
~\cite{zhang2019large} allows for the visual and language features of the subjects, objects, and relations to be adapted into a common embedding space using a visual and language embedding sub-networks. This was shown to make the model more expressive and outperform previous approaches, such as knowledge distillation~\cite{yu17iccv}, ViP-CNN \cite{LiCVPR2017}, and \cite{plummerPLCLC2017}.
\cite{OpenImages} introduced a long-tailed dataset with $600$ objects and $57$ relations. Our benchmarks focus on much larger vocabularies (see Sec. \ref{sec_datasets}).

\noindent \textbf{Long-tail Classification}
%~\cite{10.2307/2333344,10.5555/89086.89095,PIELOU1966131,10.2307/2333389,CHEN1999359,10.5555/176313.176316,salakhutdinov2011learning,zhu2014capturing,bengio2015the,liu2015deep,zhu2016we,ouyang2016factors}
Long-tail classification has been extensively studied in the literature~\cite{10.2307/2333344,10.5555/89086.89095,10.2307/2333389,salakhutdinov2011learning,zhu2014capturing,bengio2015the,liu2015deep,zhu2016we,ouyang2016factors}.
%, with the more recent approaches being inspired from \emph{metric learning}~\cite{huang2016learning,oh2016deep}, imposes~\emph{hard negative mining}~\cite{dong2017class,lin2017focal}, \emph{meta learning}~\cite{ha2016hypernetworks,wang2017learning}. 
%A variety of useful learning signals has been proposed including: lifted structure loss~\cite{oh2016deep}, range loss~\cite{zhang2017range}, and focal loss~\cite{lin2017focal}. 
Focal loss~\cite{lin2017focal} down-weights the loss assigned to well-classified examples which guides the optimizer to attend more to tail classes which are likely not well classified. % in particular has shown some promising value as it may allow the training .
% Here, our long-tail learning considers a more extreme case than the traditional imbalanced classification. Our tail classes only contain 1$\sim$20 samples while the tail classes in imbalanced classification still have 20$\sim$50 samples. More importantly, the open classes are considered in our evaluation.
In~\cite{liu2019large} the authors utilize a dynamic visual memory module and a modulated attention mechanism for generalizing over tail classes.
%proposed a dynamic meta-embedding meta learning for long-tail recognition that combines the ideas from both metric learning and meta learning. They maintain the dynamic notion by expressing an explicit visual memory model over the centroids of the classes (  allowed to change while maintaining discrimination between categories). The features are adapted with loss in addition to a memory feature, produced by an attention layer applied over the learned centroids.
In~\cite{Kang2020Decoupling}, the authors decouple the representation learning from classifier learning and show huge improvements on long-tail classification. 
%They showed that this simple training methodology outperforms state of the art approaches on long-tail classification problem.
%the authors decouple the representation learning from the classifier learning. They show that best representations are learned when the model is trained without balancing or oversampling the tail classes, while the best classifier is learned when oversampling tail classes. Since different stages of the model require different training conditions, the authors show that it is better to learn them separately instead of jointly. 
Similar to weighted CE Loss, in \cite{9157168}, the authors propose an equalization loss that blocks out gradients from affecting rare classes when training frequent classes, which improved the performance on rare classes. 
%They show that by using such a loss, they protect rare classes from being forgotten by the model due to common classes. 
In contrast, we improve the performance on tail classes by our RelMix augmentation strategy and VilHub regularizer.

\begin{figure*}[t!]
    \centering
    \includegraphics[width=.8\linewidth]{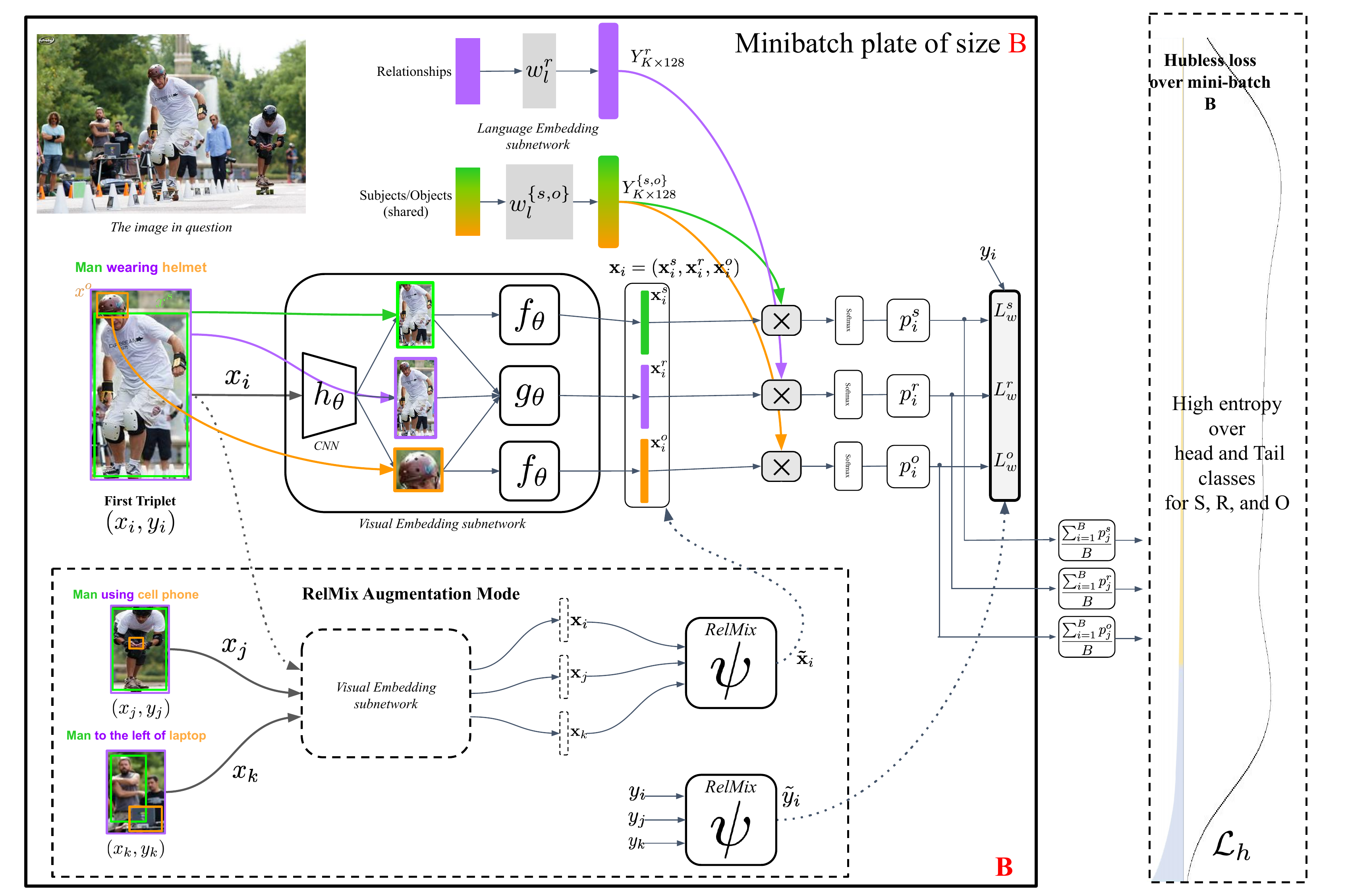}
    \caption{Approach Overview: \emph{s-r-o} triplets (extracted from the same image) are passed through the visual embedding network, the resultant embeddings are augmented by our method using a combination of features individually from subject, object, and relations. The training is then regularized with ViLub Loss and RelMix augmentation as illustrated.}%, the class probabilities are computed with the added loss term for VilHub regularizer.}
    \label{fig:approach}
\end{figure*}
\noindent \textbf{Augmentation}
There has been much work~\cite{cutmix2019, zhang2018mixup, pmlr-v97-verma19a, devries2017cutout, mancini2020towards} in recent years on using augmented data to target better generalization for classification problems. One of the better-known techniques, Mixup ~\cite{zhang2018mixup}, trains a neural network on convex combinations of pairs of examples and their labels. Manifold Mixup~\cite{pmlr-v97-verma19a} builds up on Mixup by using combination of image features rather than raw images for their augmentation technique. %This strategy can be seen as a form of data augmentation that encourages the model to behave linearly in-between training examples. They argue that this linear behaviour reduces the amount of undesirable oscillations when predicting outside the training examples. 
Cumix~\cite{mancini2020towards} proposes an extension of the Mixup technique to have combination between pairs belonging to different domains for better performance in domain generalization and zero-shot tasks.
% \noindent \textbf{Mixup}~\cite{zhang2018mixup}.
% Mixup trains a neural network on convex combinations of pairs of examples and their labels.
 % paraphrase, these sentences are taken from their paper, so paraphrase to avoid plagiarism
% \noindent \textbf{Cutout}~\cite{devries2017cutout}. In~\cite{devries2017cutout}, they proposed a form of augmentation where they randomly cutout box regions of the training images, while keeping the label as it is. They show that this simple form of augmentation greatly improves the generalization of the model.
% CutMix ~\cite{cutmix2019} is an augmentation strategy where patches from the training data are cut and pasted (i.e. the face of a cat is pasted on a dog's body), and also the ground truth is updated to reflect the ratio of patch area to original image area (i.e. Cat: 0.4, Dog: 0.6). They show that this augmentation strategy significantly improves performance over various tasks.
%Our technique is build upon this but for the visual relationship recognition tasks and consistently improves the overall performance and specially focuses on improving the tail classification performance.
% Our technique is inspired from this, but for visual relationship detection where the technique is specifically altered to focus on augmenting the lesser frequent sub-obj pairs and hence in a way, equalizing the training distribution.
We propose an augmentation technique inspired by Manifold Mixup for our LTVRD setting for a better generalization on the whole frequency band (many, med, few) of classes, with a special focus on tail classes.
% Also with our augmentation technique being model-agnostic, it can easily be integrated with any Visual Relationship Recognition models.

 %\vspace{-1mm}
% \textcolor{red}{Training using augmented data has long been effectively used as a remedy for over-fitting(e.g.,~\cite{perez2017effectiveness,zhang2018mixup,cutmix2019}), where the training data is slightly altered, not enough to change the category of the image but enough to prevent the model from memorizing the training samples. Recently, augmentation has been utilized differently, where the image is changed enough to alter the training image category, while also altering the label accordingly. This was shown to greatly improve the  decision boundaries between classes, making them more gradual and intuitive.%~\cite{cutmix2019}.
% Also the RelMix technique, being model-agnostic, can easily be integrated with any of the Visual Relationship Recognition models easily to further improve its performance on tail classes, along with an overall performance improvement
% }

% \section{Background}
% \label{sec_background}
% \input{background}
\vspace{-1mm}
\section{Approach}
\vspace{-1mm}
\label{sec_approach}
%We describe the RelMix algorithm here in detail.

%% Have some differences here listed between the cutmix and relmix here and also in the related work section.

%% Update the method section for CuMix:
% 3.1: Hubless
% 3.2: CuMix
% Update the Fig 1 & 2 to represent the new equation
% ^^This should be done asap
%% Don't cite cumix paper much so as to avoid confusion
In a visual relationship $<s,r,o>$, we define $ {\vx^{s}}, {\vx^{o}}, {\vx^{r}}$:
\begin{equation}
\small
\begin{split}
{\vx^{s}} &= f(h(x^{s}), b^s) \\
{\vx^{o}} &= f(h(x^{o}), b^o) \\
{\vx^{r}} &= g(h(x^{r}), b^r, b^s, b^o)\\
\end{split}
\label{eq_xs_xo_sr}
\end{equation}
Where $x^{s}$, $x^{r}$, and $x^{o}$ are the cropped image regions of the subject $s$, relationship $r$, and the object $o$. $b^r$, $b^s$, and $b^o$ are the corresponding bounding boxes.  ${\vx^{s}}, {\vx^{o}}, {\vx^{r}}$ are the  transformed embeddings of $x^{s}$, $x^{r}$, and $x^{o}$ respectively with corresponding labels are $y^{s}$, $y^{r}$, and $y^{o}$. $h(\theta)$ consists of the first 5 layers of VGG16, it takes the cropped image regions as input and outputs the visual features. $f(\theta)$ and $g(\theta)$ are neural networks that extract the visual embeddings from the visual features; see Fig.~\ref{fig:approach}.

% \subsection{Visiolinguistic Hubless Loss.}
\subsection{Loss Function}
\label{sec_vilhub}
% \subsection{Per-example Level Loss}

\textbf{\noindent \textbf{Per-example Loss.}} Given a set of each positive visual-language pair by $(\vx^{l}, \vy^{l})$, where $l\in\{s,r,o\}$, represented by the aforementioned neural networks, joint vision-language embeddings can be learned by a traditional triplet loss (e.g.,~\cite{kiros2014unifying,vendrov2015order,faghri2017vse++}). The triplet loss  encourages matched embeddings from the paired modalities to be closer than the mismatched ones by a margin $m$. 
% Formally, we denote a triplet $tri^l_\vy = \{ \vx^{l}, \vy^{l}, \vy^{l-}\}$ where $l \in \{ s,r,o\}$ with negatives from the language  space. 
% Omitting the superscripts $\{s,r,o\}$ for simplicity, the triplet loss $\mathcal{L}^{Tr}$ for each branch is defined as:
% \begin{equation}
% \begin{split}
% \mathcal{L}_{\vy}^{Tr} = \frac{1}{NK} \sum_{i=1}^{N} \sum_{j=1}^{K} \max [0, m + \vx_i^\mathbb{T} \vy_{ij}^- - \vx_i^\mathbb{T} \vy_i] 
% \end{split}
% \end{equation}
% where $N$ is the number of positive ROIs, $K$ is the number of negative samples \textit{per positive} ROI, $m$ is the margin between the distances of positive and negative pairs. 
The triplet loss, however, does not sense a learning signal  beyond the margin, and the trained model will not learn to  distinguish different classes enough for a classification-oriented task. To alleviate this problem, ~\cite{zhang2019large} recently studied a Softmaxed version of the triplet loss for VRR achieving state-of-the-art results. Triplet Softmax loss can be defined as follows (we drop the superscript $l\in \{s,p,o\}$ in this section for simplicity):
\begin{equation}
\small
\begin{split}
{L}(\vx_i, \vy_i) &= \frac{1}{N} \sum_{i=1}^{N} -\log  \big ( p_i= \frac{ e^{\vx_i^\mathbb{T} \vy_i}}{e^{\vx_i^{\mathbb{T}} \vy_i} + \sum_{j=1}^{K} e^{\vx_i^{\mathbb{T}} \vy_{ij}^-}} \big) \\
&= \frac{1}{N} \sum_{i=1}^{N} -\log \big ( p_i=\frac{ e^{\vx_i^\mathbb{T} \vy_i}}{ \sum_{j=1}^{K} e^{\vx_i^\mathbb{T} \vy_{j}}} \big)
\end{split}
\label{eq_trsmax}
\end{equation}
Where $N$ is the number of positive ROIs.
For each positive pair $(\vx_i, \vy_i)$ and its corresponding set of negative pairs $(\vx_i, \vy_{ij}^-)$, the  similarities between each of them is computed with dot product and then put into a softmax layer followed by multi-class logistic loss so that the similarity of positive pairs would be pushed to be $1$, and $0$ otherwise. In Eq.~\ref{eq_trsmax}, we show that triplet softmax can be simplified in a form that is very similar to MCE loss if all the other classes except the ground truth are considered negative.  We adopted a weighted version of this visiolingual loss, where we allow each class to have a weight $w_i$, this weight can be assigned higher values to less frequent classes (e.g., inverse the frequency of the object/relation class); see Eq.~\ref{eq_wtrsmax}.
\begin{equation}
\begin{split}
{L}_{w}(\vx_i, \vy_i) = \frac{1}{N} \sum_{i=1}^{N} - w_i \log \big ( p_i=\frac{ e^{\vx_i^\mathbb{T} \vy_i}}{ \sum_{j=1}^{K} e^{\vx_i^\mathbb{T} \vy_{j}}} \big )
\end{split}
\label{eq_wtrsmax}
\end{equation}
% It is worth noting that although theoretically traditional triplet loss can pushes the margin as much as possible when $m=1$, most previous works (\eg, \cite{kiros2014unifying,vendrov2015order,faghri2017vse++}) adopted a small $m$ to allow slackness during training. It is also unclear how to determine the exact value of $m$ given a specific task. We follow previous works and set $m=0.2$ in all of our experiments.
%\jun{general comments: it feels to me that you introduce a lot of related work instead of elaborating your loss functions. Shouldn't some of them be put in the related work and put more time to describe your method. It really distracts me when I read the whole story.}\\
% \textbf{\noindent \textbf{Our Final Loss XXXXXXX (Rework Hubless .}}
\textbf{\noindent \textbf{VilHub Per-minibatch Loss:}}
Recent NLP approaches like~\cite{radovanovic2010hubs,dinu2014improving,smith2017offline} observed that the accuracy of bidirectional retrieval across languages is often significantly degraded by a phenomenon called hubness, which appears when some frequent words, called hubs, get indistinguishably close to many other less represented words. 
In long-tail VRR context, these hubs are the head classes,  which are often over-predicted at the expense of tail classes. To alleviate the hubness phenomenon, we develop a vision $\&$ language hubless loss (dubbed VilHub). Our approach alleviates the long-tail problem by correcting both the language and visual representations in an end-to-end manner. The key idea of our VilHub loss $\mathcal{L}_{\mathbf{h}}$ is to encourage fair prediction over both head and tail classes in the current batch.  $\mathcal{L}_{\mathbf{h}}$ is defined as:% \jun{Describe a bit what will be your model advantage compared to other models} To our knowledge, we are the first to develop this idea into a structured visio-lingual setup to benefit the tail classes. 
\begin{equation}
\small
\begin{split}
\mathcal{L}_{\mathbf{h}} &=  \sum_{i=1}^{K} (pf(\vy_i)  - \frac{1}{K})^2,\\
pf(\vy_i)  &= \frac{1}{B} \sum_{j=1}^{B}  \frac{e^{\vx_j^\mathbb{T}  \vy_i}}{ \sum_{k=1}^{K} e^{\vx_j^\mathbb{T}  \vy_k}}\\
\end{split}
\label{eq:hubless_softmax1}
\end{equation}
Where $B$ is the mini-batch size, $K$ is the number of classes. The VilHub  loss  $\mathcal{L}_{\mathbf{h}}$ encourages all the classes (head and tail) to be equally preferred. To achieve this behavior, we define the preference of every class as $ pf(\vy_i)$ as the average probability of the class being predicted in the current minibatch of size $B$, as shown in Fig.~\ref{fig:approach}. Then, we simply encourage this marginal probability to be close to uniform (i.e., equally preferred across head and tail).% as ViLHub is marginalized over all training minibatches. % When this marginalized over the training minibatches, the model gets more predictive of . %\jun{why? is it because you guarantee that for every batch, you have a tail class and a head class?}\\

\noindent \textbf{Our Final Loss. } In conclusion, our final loss is defined as:
\begin{equation}
\small
\begin{split}
\mathcal{L} = \frac{1}{B} \sum_{i=1}^{B} {L}_{w} (\vx_i, \vy_i) +& \gamma\mathcal{L}_{\mathbf{h}}, 
% = & \alpha\frac{1}{NK} \sum_{i=1}^{N} \sum_{j=1}^{K} \max [0, m + s(\vy_i, \vx_{ij}^-) - s(\vx_i, \vy_i)] \\
% + & \beta\frac{1}{N} \sum_{i=1}^{N} -\log \frac{e^{s(\vx_i, \vy_i)}}{e^{s(\vx_i, \vy_i)} + \sum_{j=1}^{K} e^{s(\vx_i, \vy_{ij}^-)}} \\
% + & \gamma\frac{1}{NK} \sum_{i=1}^{N} \sum_{l=1}^{L} \sum_{j=1}^{K} \max [0, m + s(\vx_i, \vx_{ik}^-) - \min_{l\in\mathcal{C}(l)}s(\vx_i, \vx_l)]
% \mathcal{L}_{\mathbf{h}} =  \sum_{i=1}^{K} (pf(\vy_i)  - \frac{1}{K})^2,\\
% pf(\vy_i)  &= \frac{1}{B} \sum_{i=1}^{B}  \frac{e^{\vx_i^\mathbb{T}  \vy_j}}{ \sum_{j=1}^{K} e^{\vx_i^\mathbb{T}  \vy_j}}
\end{split}
\label{eq:hubless_softmax2}
\end{equation}
%Where $B$ is the batch size, $K$ is the number of classes, and 
where $\gamma$ is the VilHub loss weight or scale. The first term ${L}_{w}$ encourages the examples to be discriminatively classified correctly in the visual-language space. The second term $\mathcal{L}_{\mathbf{h}}$ encourages a fair prediction over head and tail classes.  

\subsection{RelMix Augmentation}
\label{sec_relmix}
We denote the input image as $x$, where there exists a visual relationship between subject $s$ and object $o$ with relationship  $r$. A visual embedding network processes the image to output three visual embeddings ($\vx^{s}, \vx^{r}, \vx^{o}$) with corresponding ground truth labels as ($y^{s}$, $y^{r}$, $y^{o}$) for subject, relationship and object.
% We also denote the corresponding ground truth labels of the relationships as $y^{s}$,   $y^{r}$, and $y^{o}$  for subject, relationship and object, respectively. Given an image, a visual embedding network processes the image to output three visual embeddings ($\vx^{s}, \vx^{r}, \vx^{o}$) for subject, relation and object respectively. 
Our RelMix algorithm augments the training data by using a combination of these features in a systematic way to help target the tail classes in the dataset.

\noindent \textbf{ RelMix.}
The goal of RelMix augmentation is to enrich training coverage of visual relationship labels by combining the extracted visual features generated in a meaningful way. During training, three triplets are selected $\vx_i = (\vx_{i}^{s}, \vx_{i}^{r}, \vx_{i}^{o}$), $\vx_j = (\vx_{j}^{s}, \vx_{j}^{r}, \vx_{j}^{o}$), $\vx_k = (\vx_{k}^{s}, \vx_{k}^{r}, \vx_{k}^{o}$)) with corresponding labels ($y_{i}=(y_{i}^{s}, y_{i}^{r}, y_{i}^{o}), y_{j}=(y_{j}^{s}, y_{j}^{r}, y_{j}^{o}), y_{k}=(y_{k}^{s}, y_{k}^{r}, y_{k}^{o})$). We sample these  triplets such that  ($y_{i}$ and $y_{j}$) belong to the tail/medium classes (high class frequency domain) while the triplet ($y_{k}$) belongs to the tail class ( low class frequency domain). %Let the frequency of the relationship ground truth class ($y_{i}^{r}, y_{j}^{r}, y_{k}^{r}$) in these triplets be ($f_{i}, f_{j}, f_{k}$). Then these triplets are selected in such a way that ($f_{i} < f_{j} < f_{k}$).
Then inspired from  existing augmentation strategies (e.g.,~\cite{pmlr-v97-verma19a}), we combine the the input embeddings and corresponding predictions as in Eq.~\ref{eq_relmix}. 
\begin{equation}
\begin{aligned}
    \tilde{\vx} = \psi (\vx_{i}^l, \vx_{j}^l, \vx_{k}^l) = \lambda \vx_{i}^l + (1- \lambda)(\alpha \vx_{j}^l + (1 - \alpha) \vx_{k}^l) \\
   \tilde{y} =   \psi (y_{i}^l, y_{j}^l, y_{k}^l) = \lambda y_{i}^l + (1- \lambda)(\alpha y_{j}^l + (1 - \alpha) y_{k}^l)
\end{aligned}
\label{eq_relmix}
\end{equation}
Where $l\in \{s,r,o\}$,  $\alpha$ is sampled from Bernoulli distribution with probability 0.5, and $\lambda\in [0,1]$. %\textcolor{red}{Aniket: some others were also seen like 0.6-0.7 gave good results}, and.
%%Here this combination is done for subject, object and relationship individually with the same parameters.
This allows our method to focus on long-tail classes more since the three triplets are chosen in a way to have the augmented features resemble closely to tail ones. The frequent classes are sampled in one out of three triplets to maintain representation quality while the aforementioned VilHub loss encourages fair prediction over all head and tail classes; (see Fig.\ref{fig:approach}).

While RelMix is inspired from Manifold mixup~\cite{pmlr-v97-verma19a}, with its augmentation being done at the feature representation level, there are some key differences, \textbf{(1).} Manifold mixup operates on object recognition setting ( one object / image), compared to VRD setting where mini-batches consist of scenes; each having multiple objects and structured relations. \textbf{(2).} Manifold Mixup extracts the (object, label) pairs from many different images, which is not so practical in the VRD setting. We instead extract augmentation tuples of s-r-o triplets from the same scene, so the training efficiency is not significantly hurt.  Concretely, we augment the most frequent classes with the least frequent ones in the same scene and vice versa. Further comparison between RelMix and Manifold Mixup is taken up in Section~\ref{sec_ablation}.
% To have a more concrete comparison between RelMix and Manifold Mixup, we further take this up in Section~\ref{sec_ablation}.

% \textcolor{blue}{The technique is partly inspired from Manifold Mixup \cite{pmlr-v97-verma19a} with its augmentation being done at the feature representation level. However there are some key differences, \textbf{(1).} Manifold mixup operates on object recognition setting (mainly one object / image), compared to VRD setting where mini-batches consist of scenes; each having multiple objects and structured relations. \textbf{(2).} Manifold Mixup extracts the (object, label) pairs from many different images which is not so practical in VRD setting. We instead extract augmentation tuples of s-r-o triplets from the same scene, so the training efficiency is not significantly hurt. More concretely, we augment the most frequent classes with the least frequent ones in the same scene and vice versa. \textbf{(3).} As \textbf{\ReviewerA} pointed out, we mixup three samples of s-r-o triplets instead of two. Our intuition is to consider head and non-frequent tail classes as two domains. For an example $i$ from head/tail, we sample $j$ from the other domain tail/head, and $k$ from the same domain head/tail; see Eq. 6. We found this design choice better regulates the learning representation to balance between head and tail. When we tried 2 samples instead of 3 on LTVRD, the performance degrades by 1.2\% on the few-shot classes for sbj/obj category on GQA-LT, while  maintaining the performance on more frequent classes}

\noindent \textbf{Augmentation Loss.} Once $\tilde{\vx}$ and $\tilde{y}$ is computed, they are then fed to our final loss, defined in Eq.~\ref{eq:hubless_softmax2}.

\section{Experiments}
\label{sec:experiments}

\subsection{Datasets and Comparison Models} 
\label{sec_datasets}
We present experiments on two LTVRR benchmarks that we built on top of Visual Genome~\cite{krishna2017visual,zhang2019large} and GQA dataset~\cite{hudson2019gqa} (also based on VG). Both datasets naturally have a long-tail distribution yet only high  frequency  subjects, relations, and objects are mainly used in the literature. 
% Make the following changes:
% 1. Change the language in the datasets section from the LTVRR papers.
% 2. Change the figure a bit so make it one-column figure and also reduce its size and add it to page 1
% 3.

\noindent \textbf{GQA-LT}. We used the visual relationship notations provided with the GQA dataset~\cite{hudson2019gqa}. %which we found following a long-tail distribution. 
The main filtration we applied to GQA data was to remove the objects that did not belong to a subject-relation-object triplet. %, because we  focus on  relationships, not object detection.
The resulting benchmark has $72,580$ training images, $2,573$ validation images, and  $7,722$ test images;
with $1,703$ objects and $310$ relations. We call this version GQA-LT. Most frequent object and relation has $374,282$ and $1,692,068$ examples, and the least frequent are $1$ and $2$, respectively.%`We split the test set into $7,722$ test images and $2,573$ validation images.
%and $10,295$ testing images,

\noindent \textbf{VG8K-LT} We used the latest version of Visual Genome (VG v1.4) \cite{krishna2017visual} that contains $108,077$ images with $21$ relationships on average per image. We used the data split in ~\cite{zhang2019large} which has  $103,077$ training images and $5,000$ testing images following~\cite{densecap} and used the class labels that have corresponding word embeddings~\cite{mikolov2013distributed}.  

We selected the most frequent $5,330$ object classes out of the original $53,304$ and $2,000$ relationships out of the original $29,086$ to make a cleaner version of VG80K (noisy).
The resulting dataset has $97,623$ training images,  $1,999$ validation images, and $4,860$ testing images.  
%We split the training set into $97,623$ training images and $1,999$ validation images.
%The resulting dataset has $99,622$ training images, and $4,860$ testing images. We split the training set into $97,623$ training images and $1,999$ validation images.
After the filtration the least frequent object and relation classes have $14$ and $18$ examples, and the most frequent are $196,944$ and $618,687$, respectively, meaning the distribution is very long-tailed. We call this version VG8K-LT.

\begin{table}[t!]
% \parbox{.5\linewidth}{
\centering
\caption{\label{tab:gqa_many_median_few_pc_syn} Average per-class accuracy on GQA-LT}
\scalebox{0.6}{
\begin{tabular}{lccccccccc}
 \hline \hline
 \multicolumn{1}{c}{} & \multicolumn{4}{c}{Subject/Object} & \multicolumn{4}{c}{Relation} \\
Loss & many &  medium &  few &  all &  many &  medium &  few &  all \\
 & 86 & 255 & 1,362 &  1,703 & 16 & 46 & 248 & 300\\
\hline
CE \cite{zhang2019large} &68.3 & 37.0 & 6.9 & 14.5 & 62.6 & 15.5 & 6.8 & 11.0\\
CE + VilHub & 68.6 & \textbf{44.0} & \textbf{10.3} & \textbf{18.3} & \textbf{63.6} & \textbf{17.6} & 7.2 & 11.7\\
%LSVRU & RelMix & 68.2 & 37.7 & 9.3 & 16.5 & 62.3 & 16.0 & 6.6 & 10.9\\
CE + VilHub + RelMix & \textbf{68.8} & 42.1 & 10.1 & 18.1 & 63.4 & 14.9 & \textbf{8.0} & \textbf{11.9}\\
\hline
Focal Loss \cite{lin2017focal} & 68.2 & 39.2 & 7.5 & 15.3 & 60.4 & \textbf{15.7} & \textbf{7.7} & \textbf{11.6} \\
Focal Loss + VilHub & \textbf{69.0} & \textbf{43.4} & \textbf{9.5} & \textbf{17.5} & \textbf{63.1} & 14.2 & 7.5 & 11.4 \\
\hline
OLTR \cite{liu2019large} & 68.2 & 37.2 & 7.0 & 14.6 & 62.3 & 15.8 & 6.6 & 10.8 \\
OLTR + VilHub & \textbf{69.1} & \textbf{38.7} & \textbf{7.6} & \textbf{15.2} & \textbf{63.0} & \textbf{16.8} & \textbf{7.3} & \textbf{11.2} \\
\hline
DCPL \cite{Kang2020Decoupling} & 64.0 & 35.3 & 6.4 & 13.7 & \textbf{61.4} & 23.6 & \textbf{7.6} & \textbf{12.7} \\
DCPL + VilHub & 63.5 & 39.8 & 7.5 & 15.2 & 58.6 & \textbf{26.1} & 7.0 & 12.5 \\
DCPL + VilHub + RelMix & \textbf{65.7} & \textbf{40} & \textbf{7.8} & \textbf{15.4} & 58.9 & 25.7 & 6.8 & 12.3 \\
\hline
EQL \cite{9157168} & 68.9 & 43.7 & 10.0 & 18.0 & 63.5 & 15.0 & 8.2 & 12.1 \\
EQL + VilHub & 67.7 & 43.9 & 11.1 & 18.7 & 62.8 & 15.8 & 8.9 & 12.6 \\
EQL + VilHub + RelMix & \textbf{69.1} & \textbf{44.3} & \textbf{11.3} & \textbf{18.8} & \textbf{64.1} & \textbf{16.4} & \textbf{9.2} & \textbf{12.8} \\
\hline
WCE & \textbf{53.4} & 42.0 & 14.0 & 20.2 & \textbf{53.4} & 35.1 & 15.7 & 20.5 \\
WCE + VilHub & 52.0 & 44.6 & \textbf{16.0} & \textbf{22.1} & 53.1 & 39.0 & \textbf{15.8} & \textbf{21.2} \\
WCE + VilHub + RelMix & 52.7 & \textbf{45.2} & 15.7 & 22 & 55.1 & \textbf{39.3} & 15.7 & 21.1 \\
\hline
\end{tabular}
}
% }
\end{table}

\begin{table}
% \parbox{.5\linewidth}{
\centering
\caption{\label{tab:vg8k_many_median_few_pc_syn} Average per-class accuracy on VG8K-LT}
\scalebox{0.6}{
\begin{tabular}{lccccccccc}
 \hline \hline
 \multicolumn{1}{c}{} & \multicolumn{4}{c}{Subject/Object} & \multicolumn{4}{c}{Relation} \\
Loss & many &  medium &  few &  all &  many &  medium &  few &  all \\
& 267 & 799 & 4,264 & 5,330 & 100 & 300 & 1,600 & 2,000\\
\hline
CE \cite{zhang2019large} & 57.3 & 11.1 & 8.5 & 11.4 & 22.2 & 15.5 & 12.6 & 13.5 \\
CE + VilHub & \textbf{61.6} & \textbf{20.3} & 10.1 & \textbf{14.2} & \textbf{27.5} & \textbf{17.4} & \textbf{14.6} & \textbf{15.7} \\
%LSVRU & RelMix & 56.9 & 12.4 & 10.3 & 13.0 & 22.7 & 15.6 & 12.6 & 13.6 \\
CE + VilHub + RelMix & 59.5 & 15.1 & \textbf{10.4} & 13.6 & 24.5 & 16.5 & 14.4 & 15.4 \\
\hline
Focal Loss \cite{lin2017focal} & 58.1 & 13.9 & 8.9 & 12.1 & 24.5 & \textbf{16.2} & 13.7 & 14.7 \\
Focal Loss + VilHub & \textbf{60.5} & \textbf{16.7} & \textbf{9.2} & \textbf{12.9} & \textbf{26.7} & 15.7 & \textbf{13.9} & \textbf{14.8} \\
\hline
OLTR \cite{liu2019large} & 56.8 & 12.0 & 9.6 & 12.3 & 22.5 & 15.6 & 12.6 & 13.6 \\
OLTR + VilHub & \textbf{60.4} & \textbf{15.1} & \textbf{9.8} & \textbf{13.1} & \textbf{27.8} & \textbf{16.4} & \textbf{14.4} & \textbf{15.4} \\
\hline
DCPL \cite{Kang2020Decoupling} & 53.8 & 5.9 & 7.9 & 9.9 & 34.4 & 15.4 & \textbf{12.9} & \textbf{14.4} \\
DCPL + VilHub & 56.4 & 7.0 & 8.2 & 10.4 & 35.2 & 15.3 & 12.8 & 14.3 \\
DCPL + VilHub + RelMix & \textbf{57.6} & \textbf{7.4} & \textbf{8.3} & \textbf{10.5} & \textbf{35.9} & \textbf{15.5} & 12.8 & 14.3 \\
\hline
EQL \cite{9157168} & 56.9 & 12.1 & 10.0 & 12.7 & 22.6 & 15.6 & 12.6 & 13.6 \\
EQL + VilHub & 60.3 & 15.0 & 10.2 & 13.4 & 27.9 & 16.5 & \textbf{14.4} & 15.4 \\
EQL + VilHub + RelMix & \textbf{62.1} & \textbf{15.1} & \textbf{10.4} & \textbf{13.6} & \textbf{29.3} & \textbf{16.9} & 14.3 & \textbf{15.5} \\
\hline
WCE & 52.8 & 27.2 & 10.8 & 14.5 & 35.5 & 24.7 & 15.2 & 17.2 \\
WCE + VilHub & 52.0 & \textbf{27.9} & \textbf{11.1} & \textbf{14.8} & 35.2 & 24.6 & \textbf{15.3} & \textbf{17.2} \\
WCE + VilHub + RelMix & \textbf{54.2} & 26.7 & 10.3 & 14.1 & \textbf{36.8} & \textbf{25.3} & 14.2 & 16.5 \\
\hline
\end{tabular}
}
% }
\end{table}

 \textbf{Comparison Models.} We compare our method with several state-of-the-art
approaches that focus on the long-tail~\cite{zhang2019large,lin2017focal,Kang2020Decoupling, 9157168}. For fair comparisons, we  use the same backbone neural network in~\cite{zhang2019large} with all approaches; ~\cite{zhang2019large} is based on VGG16 architecture~\cite{simonyan2014very} .\\
\noindent \textbf{LSVRU}~\cite{zhang2019large}: this is the base  visio-lingual model with structured visual encoder. \\
\noindent \textbf{Focal Loss} (FL)~\cite{lin2017focal}: A  loss used in object detection setting to alleviate the  long-tail problem. We integrated FL with LSVRU on each of $s/o,r$ classification heads.

%tried using the same model in~\cite{zhang2019large} but add focal loss (with varying gamma values) to see how much it would help performance on the tail.\\
\noindent \textbf{Weighted Cross Entropy (WCE)}: We use a weighted version of cross-entropy loss. The weight is based on the inverse class frequency, which gives a large weight to rare classes and a small weight to common classes. 
%We integrated WCE with LSVRU on each of $s/o,r$ classification heads.   %but use a standard weighted cross-entropy instead of standard cross-entropy.\\

% \noindent \textbf{Fully Connected (FC)}: To show how much the language guidance helps the other models,  we remove the language guidance network of~\cite{zhang2019large} and replace it with 1 fully connected classification layer. 
%We use this as a baseline \\
% models~\cite{Kang2020Decoupling} against our work.\\ 

\noindent \textbf{Decoupling (DCPL)}~\cite{Kang2020Decoupling}: This is a state-of-the-art model in long-tail classification that is based on decoupling representation learning phase from classifier learning phase. We applied DCPL in our LTVRR setup similarly.  

\noindent \textbf{OLTR}~\cite{liu2019large}:  We implemented the visual memory module augmented with modulated attention by~\cite{liu2019large} into our LTVRR task using ~\cite{zhang2019large} as a  backbone model.   

\noindent \textbf{EQL}~\cite{9157168}: The equalization loss  protects the learning of tail  classes from being at a disadvantage during the network parameter updating ~\cite{9157168}.\\
%\noindent \textbf{CutMix}~\cite{cutmix2019}: This model uses the augmentation strategy from~\cite{cutmix2019}
\noindent \textbf{Visio-Lingual Hubless (VilHub)}: Models using our hubless regularizer explained in section~\ref{sec_vilhub}.\\
\noindent \textbf{RelMix}: This is the model using our proposed augmentation strategy explained in section~\ref{sec_relmix}.

\begin{figure*}[t!]
    \centering
    \includegraphics[width=1.0\linewidth]{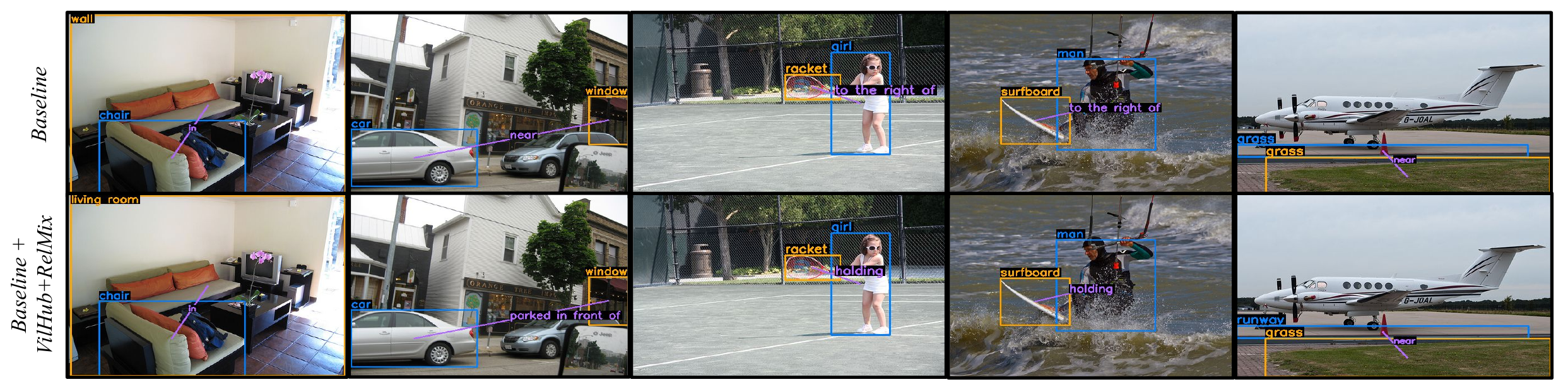}
    \caption{Qualitative examples for our model. In all of these, we see our model preferring the better choice (mostly long-tail) for sbj/obj/rel prediction.}%, the class probabilities are computed with the added loss term for VilHub regularizer.}
    \label{demonstrations}
\end{figure*}

\noindent \textbf{Metrics.} \textit{Average per-class accuracy} The main metric used in the tables is the \emph{average per-class accuracy}, which is the accuracy of each class calculated separately, then averaged. The average per-class accuracy is a commonly used metric in the long-tail literature~\cite{Kang2020Decoupling,9157168,liu2019large}.

% \textbf{Soft Average Precision} This is our new proposed metric. We propose to use a soft average precision metric,  giving partial credit to answers that are close to the ground truth,  based on six wordnet~\cite{miller1995wordnet} metrics that have been used in the literature~\footnote{reference to the implementation we used of these six functions  \url{http://www.nltk.org/howto/wordnet.html}}: Leacock-Chodorow Similarity(LCH), Wu-Palmer Similarity(WUP), Resnik (RES), Lin (LIN), Jiang-Conrath (JCN), and path similarity (PATH). In addition, we also used two word2vec metrics trained on Visual Genome (W2V-VG) and Google News (W2V-GN)~\cite{mikolov2013distributed}. For each test example, we start by producing the top 250 closest predictions and store the corresponding confidence. Then, we use each of these metrics to retrieve the top $T$ closest labels in the list of 250 predicted labels according to the chosen metric and mark it as relevant (i.e., label 1) to the given example in addition to the ground truth. We mark the remaining labels as irrelevant  (label 0). After that, we report the Average Precision (AP) at 1, and 10. This analysis tool measures the capability of a system to bring classes that are semantically close to the ground truth up in the rank of predictions.

\noindent \textbf{Many, Medium, Few splits}: We report the results on the subject, object, and relation separately of an $<$S,R,O$>$ triplet on GQA-LT and VG8K-LT datasets. We evaluate the models using the average per-class accuracy across several frequency bands chosen based on frequency percentiles for GQA-LT: \emph{many}: top 5\% frequent classes, 86 classes for S/O, and 16 classes for R. \emph{medium}: the middle 15\%, 255 classes for S/O, and 46 classes for R. \emph{few}: the least frequent 80\% of classes, 1362 classes for S/O, and 248 classes for R. VG8K-LT is split similarly; see supplementary. 

\subsection{Quantitative Results}
Table~\ref{tab:gqa_many_median_few_pc_syn} shows that adding VilHub loss (w/ and w/o RelMix) to any of the compared models consistently improve their performance on the \emph{medium} and \emph{few} classes categories. While VilHub alone can improve almost all the models in \emph{med} and \emph{few} categories, we also see the addition of RelMix further improving this in all categories for sbj/obj classification. A similar trend can also be seen when evaluating these models on VG8K-LT dataset, as seen in Table~\ref{tab:vg8k_many_median_few_pc_syn}.

Comparing LSVRU with and without VilHub in Table~\ref{tab:gqa_many_median_few_pc_syn}, VilHub loss improved the performance $\approx$7.0\% for sbj/obj \emph{medium} category, $\approx$3.4\% for the sbj/obj \emph{few} category, and $\approx$2.1\% for the relation \emph{medium} category. In this case, we also see an improvement of $\approx$1.2\% on relation \emph{few} category when combining VilHub with RelMix. We also see a consistent improvement over the whole band (\emph{many}, \emph{med}, \emph{few}) for sbj/obj and rel when using EQL~\cite{9157168} baseline in combination with VilHub and RelMix with a substantial gain of $\approx$6.7\% in the \emph{med} category for relation. A similar improvement in performance with the addition of VilHub (w/ and w/o RelMix) can also be seen in the Decoupling~\cite{Kang2020Decoupling}, OLTR~\cite{liu2019large} and Focal Loss~\cite{lin2017focal} baselines. Comparing WCE and WCE + VilHub in sbj/obj branch, we can see that adding the VilHub loss improved $\approx$2\% in the \emph{few} category, and with the addition of RelMix improved $\approx$3.2\% in the \emph{med} category. While similar behavior can be seen in VG8K-LT evaluation (Table~\ref{tab:vg8k_many_median_few_pc_syn}), we can see that the improvement on GQA-LT is more apparent than on VG8K-LT, since VG8K-LT dataset is more challenging and has more than 5 times the number of objects and more than 7 times the number of relationships compared to GQA-LT. With the model agnostic nature of VilHub+RelMix, they can be easily be integrated on top of existing VRR models to improve their performance, especially on the \emph{med} and \emph{tail} categories. Some of the qualitative results can be seen in Figure~\ref{demonstrations}.

\subsection{Ablation}
\label{sec_ablation}

\begin{table}
\caption{Ablation Study for RelMix and VilHub.}
\centering
\small
\setlength{\tabcolsep}{2.5pt}

\scalebox{0.8}{
\begin{tabular}{lcccccccc}
\toprule
 {} & \multicolumn{4}{c}{Subject/Object} & \multicolumn{4}{c}{Relation} \\
Model &  many &  med &  few &  all &  many &  med &  few &  all \\
\midrule
LSVRU~\cite{zhang2019large} & 68.3 & 37.0 & 6.9 & 14.5 & 62.6 & 15.5 & 6.8 & 11.0 \\

\midrule

LSVRU + Manifold Mixup~\cite{pmlr-v97-verma19a} & 68 & 37.5 & 7.5 & 15.1 & 62.4 & 15.7 & 6.8 & 11 \\
LSVRU + RelMix & 68.2 & 37.7 & 9.3 & 16.5 & 62.6 & 16.0 & 6.9 & 11.1 \\

\midrule

LSVRU - Lang. & 68.7 & 26.0 & 5.2 & 11.5 & 49.9 & 9.0 & 5.8 & 8.5 \\
LSVRU - Lang. + VilHub & 69.7 & 31.4 & 5.6 & 12.7 & 54.1 & 8.7 & 5.4 & 8.4 \\

\midrule

LSVRU + VilHub(1k) & 68.3 & 37.1 & 7.0 & 14.6 & 63.5 & 16.3 & 6.8 & 11.2 \\
LSVRU + VilHub(5k) & 68.4 & 38.6 & 7.4 & 15.2 & 63.6 & 17.6 & 7.2 & 11.7 \\
LSVRU + VilHub(10k) & 68.6 & 39.7 & 8.0 & 15.8 & 63.6 & 17.3 & 7.2 & 11.6 \\
LSVRU + VilHub(20k) & 68.7 & 41.0 & 8.4 & 16.3 & 63.5 & 16.5 & 7.1 & 11.4 \\

\bottomrule
% also do some ablations on scene input  & & & & & & & & \\
\\
\end{tabular}}
\label{ablation}
\end{table}

We perform ablations to better understand the influence of the proposed VilHub regularizer and RelMix augmentation; see  Table~\ref{ablation}. The entry \emph{LSVRU + Manifold Mixup} represents an adapted version of Manifold Mixup~\cite{pmlr-v97-verma19a} in our setting. We can observe a significant performance gap between the said baseline and RelMix, especially in the \emph{few} categories of sbj/obj, where the gap is $\approx$1.8\%. 
Further ablations of RelMix augmentation can be found in supp.

The \emph{LSVRU - Lang.} is our backbone model (LSVRU) without the language guidance network of~\cite{zhang2019large}, where we replace it with one FC classification layer.  Table~\ref{ablation} shows a drop in performance when removing the language guidance. We also show that VilHub loss worsens the tail relations performance when applied without language guidance.

\begin{figure}
  \centering
%   \vspace{-2mm}
\begin{subfigure}{.23\textwidth}
%   \centering
  \includegraphics[width=\linewidth]{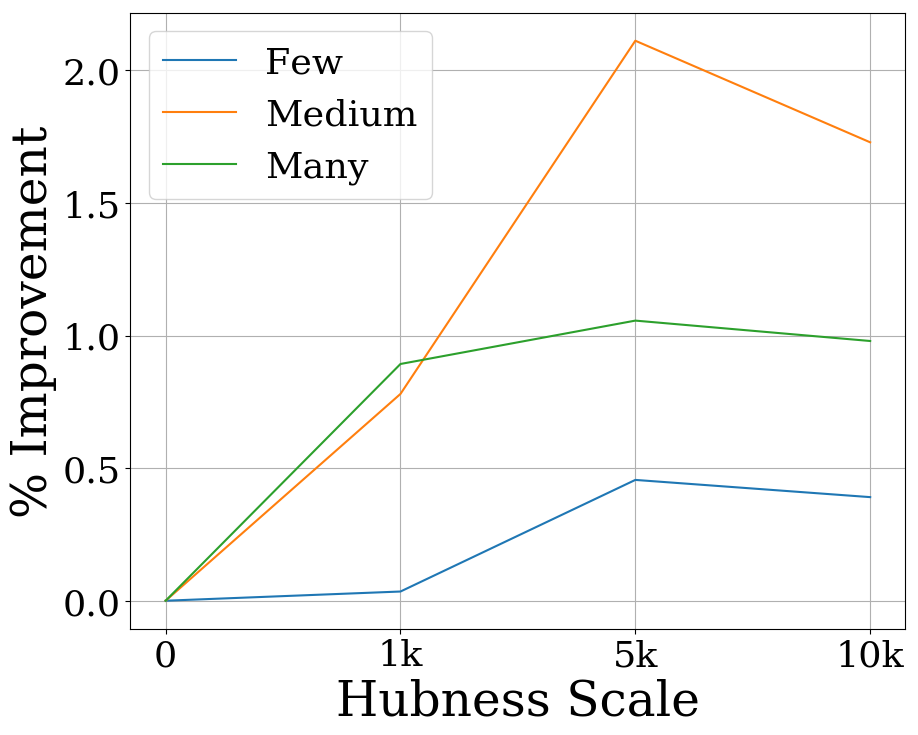}
  \caption{GQA-LT}
  \label{fig:gqa_hub}
\end{subfigure}
\begin{subfigure}{.23\textwidth}
%   \centering
  \includegraphics[width=\linewidth]{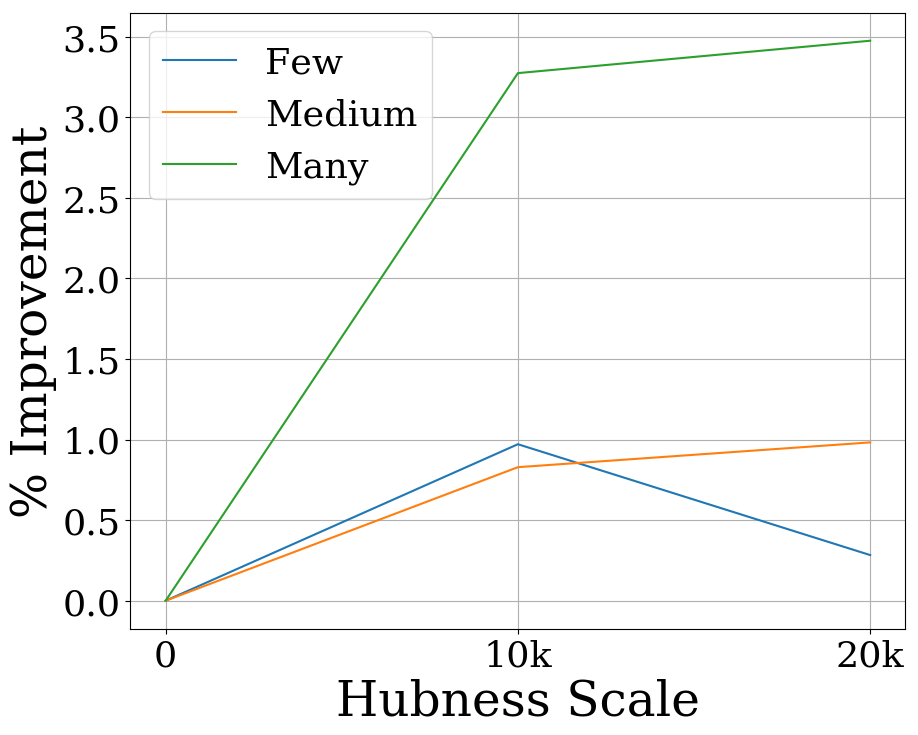}
  \caption{VG-LT}
  \label{fig:vg_hub}
\end{subfigure}
\caption{\textbf{Relationships performance change on the many, medium, few shots classes as we increase the VilHub scale}. The performance improves on medium and few (tail) classes up to a certain point and then declines.}

\label{fig:trends}
\end{figure}

We further analyze the effect of changing the scale value $\gamma$ (from Eq.~\ref{eq:hubless_softmax2}) of the VilHub regularizer (e.g., 1k, 5k, 10k, 20k). Fig~\ref{fig:trends} shows that the performance increases on each medium and few classes as we increase the VilHub scale up to an ideal value and then tends to drop. We observe that the ideal VilHub scale value tends to be higher for subjects and objects than for relationships.

\subsection{Further Analysis}
% comparisons figures
\begin{figure*}[h!]
\begin{subfigure}{.5\linewidth}
\centering
\includegraphics[width=\linewidth]{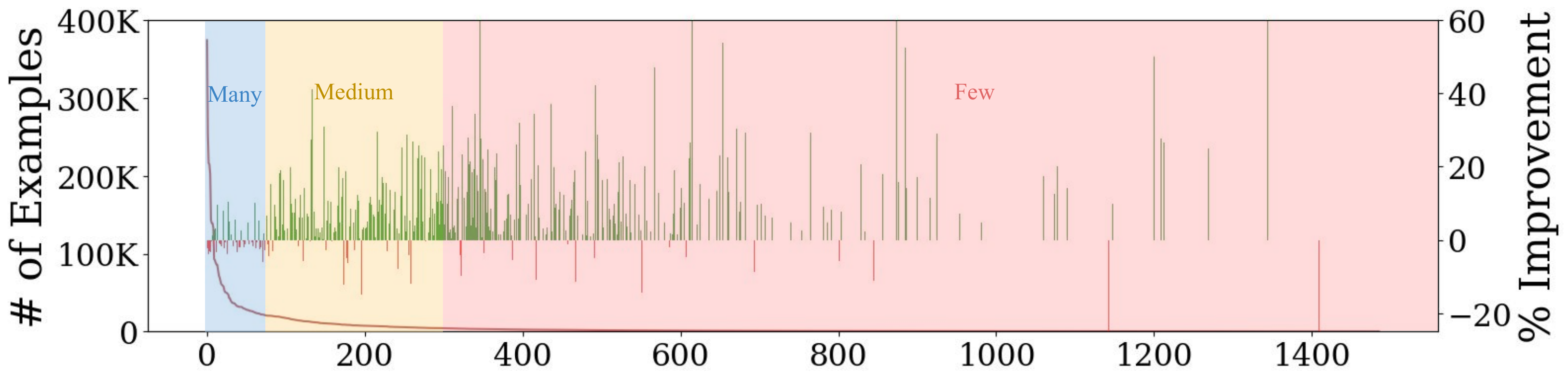}
\end{subfigure}
\begin{subfigure}{0.5\linewidth}
\centering
\includegraphics[width=\linewidth]{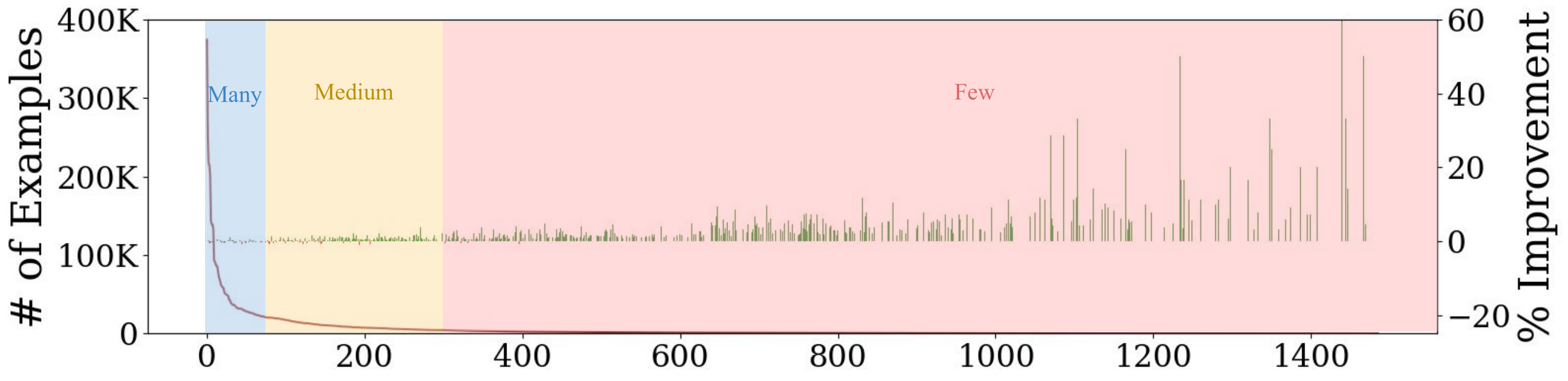}
\end{subfigure}
% \caption{Comparisons of subject/object (upper) and relations (lower) accuracy between LSVRU model with and without ViLHub on GQA-LT dataset. The performance for all classes is sorted by frequency. The distribution of classes is shown in the background (in brown). The green are improvement, red bars are worsening, no bars indicate no change. The improvement on the y-axis is the absolute improvement of the classes in percentage accuracy.}
% \label{fig:comps1}
% \end{figure}

% \begin{figure}[t]
\begin{subfigure}{0.5\linewidth}
\centering
\includegraphics[width=\linewidth]{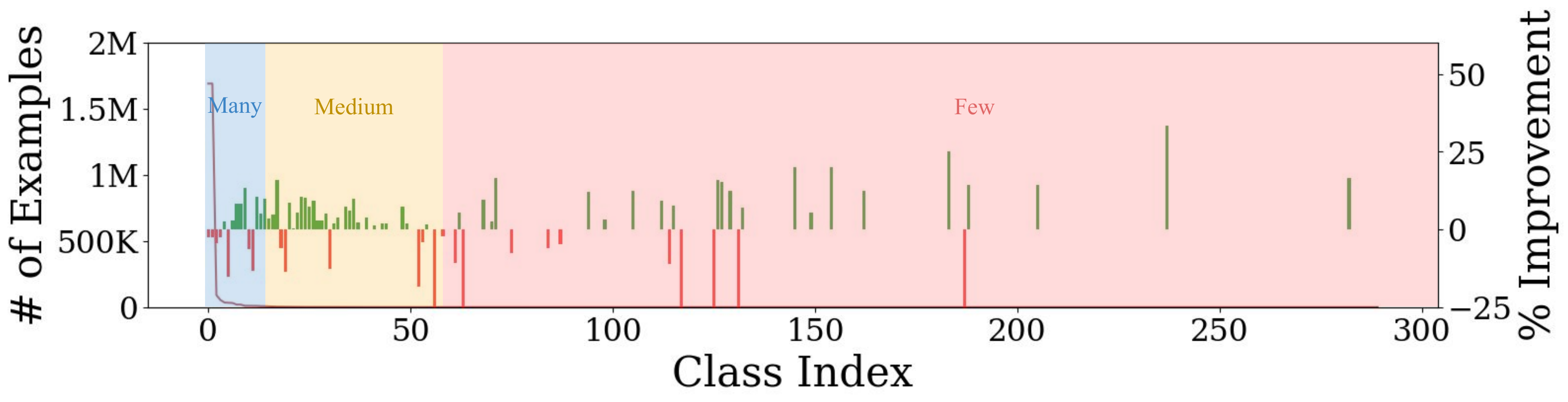}
\end{subfigure}
\begin{subfigure}{0.5\linewidth}
\centering
\includegraphics[width=\linewidth]{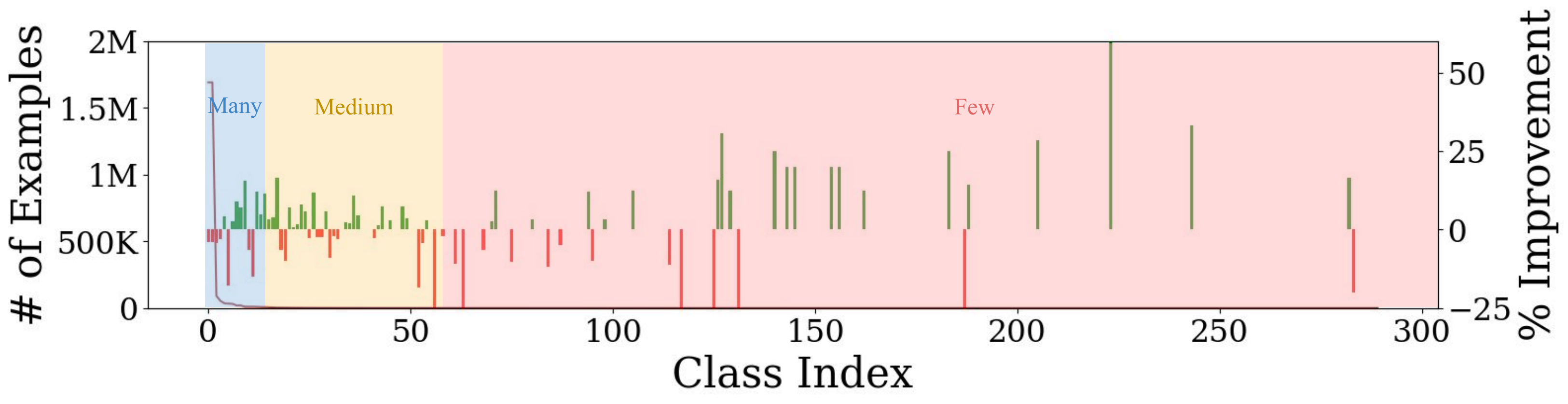}
\end{subfigure}
% \caption{Comparison of LSVRU model and LSVRU + RelMix on GQA-LT dataset for sbj/obj (top) and rel (bottom). Note how consistently Relmix improves the results on the subjects/objects (top)}
\caption{Comparisons of subject/object (upper) and relations (lower) accuracy between LSVRU model with and without VilHub (left) and LSVRU model and LSVRU + RelMix (right) on GQA-LT dataset. The performance for all classes is sorted by frequency. The distribution of classes is shown in the background (in brown). The green are improvement, red bars are worsening, no bars indicate no change. The improvement on the y-axis is the absolute improvement of the classes in percentage accuracy. Note the consistent improvements across tail classes due to VilHub and RelMix}
\label{fig:comps}
\end{figure*}

We analyze in more depth the results in Table~\ref{tab:gqa_many_median_few_pc_syn} to  better understand the causes of improvements.
Figure~\ref{fig:comps} shows head-to-head comparison between the LSVRU model~\cite{zhang2019large} with and without the VilHub regularizer.
For the subjects and objects in Figure~\ref{fig:comps} (top left), adding the VilHub loss improved 415 classes out of 1703 and only worsened 79 classes. For the relationships Figure~\ref{fig:comps} (bottom left), adding the VilHub improved 56 classes out of 310 and worsened around 24 classes.
The figures show that most of the improvement is over \emph{medium} and \emph{few} shot classes (tail). This demonstrates how adding the VilHub loss improves the performances of the classes across the frequency spectrum. On average, we see that VilHub improves many more classes than it worsens and improves more classes on the tail. This confirms our point that adding the VilHub loss as a regularizer in long-tail problems pushes the models to learn classes across the spectrum, and prevents the models from solely focusing on improving the head classes. 
% Apart from that, in Table~\ref{tab:vg8k_many_median_few_pc_syn}, we see noticeable performance improvement in the emph{many} category aside of the emph{med}/emph{few} when VilHub is applied. which can be attributed to the fact that the loss encourages visual classifiers to be more predictive of tail classes while being accurate on the head. This predictive learning signal helps better leverage examples of tail  classes  contrastively  against  head  classes  rather than being ignored. We believe that this enriched contrastive learning of tail classes helps the learning representation of head classes be more discriminative.

Figure~\ref{fig:comps} (right) shows how adding the Relmix augmentation improves the results across the classes' spectrum. Figure~\ref{fig:comps} (top right) shows how consistently RelMix improves the results on the tail end of the distribution (improved around 700 out of 1703 and worsened only 92 classes). Figure~\ref{fig:comps} (bottom right) shows an overall improvement over the classes with more focus on the tail. This shows our augmentation method's potential and motivates further research into replicating the consistent improvement on the subjects/objects (Figure~\ref{fig:comps} top right) to relationships.

In Table~\ref{tab:vg8k_many_median_few_pc_syn}, we see noticeable performance improvement in the \emph{many} category aside from the \emph{med}/\emph{few} when VilHub is applied. While this result seems surprising at first, we note that by design, our loss encourages the visual classifiers to be more predictive of tail classes while also being accurate on the head. This predictive learning signal helps better leverage tail classes examples  contrastively  against  head  classes  rather than being ignored. We believe that this enriched contrastive learning of tail classes helps the learning representation of head classes be more discriminative.

% Apart from that, in Table 2, we see noticeable performance improvement in the many category aside of the med/few when VilHub is applied. While this result seems surprising at first, we note that by design our loss encourages the visual classifiers to be more predictive of tail classes while also being accurate on the head. Specifically, instead of ignoring the tail classes, our loss enables generalization/learning on the tail; which in turn, we believe can act as an extra source of labelled data which can help also the head.

We also report our model's performance using the standard per-example accuracy in Table~\ref{per_example_table}, showing that our proposed models can improve the overall accuracy. However, we may get 90\% with the standard accuracy metric if we correctly predict only the top frequent few classes, mostly ignoring the tail. That's why per-class accuracy is adopted in LTOR literature~\cite{Kang2020Decoupling, 9157168, liu2019large}.

% per-example table
\begin{table}
\centering
\small
\setlength{\tabcolsep}{2.5pt}
\caption{Average per-example accuracy on GQA-LT}
\label{per_example_table}
\vspace{-2mm}
\scalebox{0.8}{
\begin{tabular}{lcccccccc}
\toprule
 {} & \multicolumn{2}{c}{per-example accuracy} \\
Model &  sbj/obj & rel \\
\midrule
LSVRU~\cite{zhang2019large} & 51.9 & 94.8 \\
LSVRU + VilHub & 53.9 & 91.2 \\
LSVRU + VilHub + RelMix & 53.5 & 91.0 \\

\midrule

EQL~\cite{9157168} & 51.13 & 93.85 \\
EQL + VilHub &  52.80 & 92.23 \\
EQL + VilHub + RelMix & 52.74 & 92.58 \\

\midrule

WCE & 37.6 & 72.6 \\
WCE + VilHub & 40.0 & 74.3 \\
WCE + VilHub + RelMix & 47.3 & 77.6 \\

\bottomrule
% also do some ablations on scene input  & & & & & & & & \\
\end{tabular}}
\end{table}

\begin{figure*}[h!]
\begin{subfigure}{\textwidth}
  \centering
  \includegraphics[width=.7\linewidth]{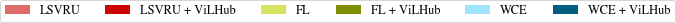}
\end{subfigure}
\begin{subfigure}{.19\textwidth}
%   \centering
  \includegraphics[width=\linewidth]{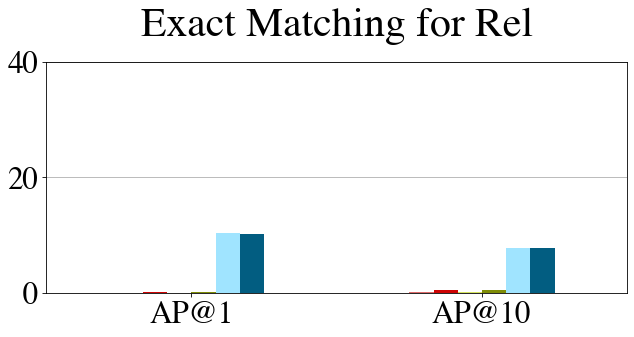}
%   \caption{}
  \label{fig:avg_prec1}
\end{subfigure}%
\begin{subfigure}{.19\textwidth}
%   \centering
  \includegraphics[width=\linewidth]{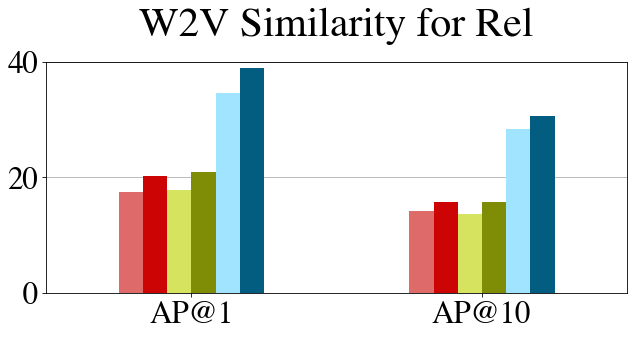}
%   \caption{}
  \label{fig:avg_prec2}
\end{subfigure}
\begin{subfigure}{.19\textwidth}
%   \centering
  \includegraphics[width=\linewidth]{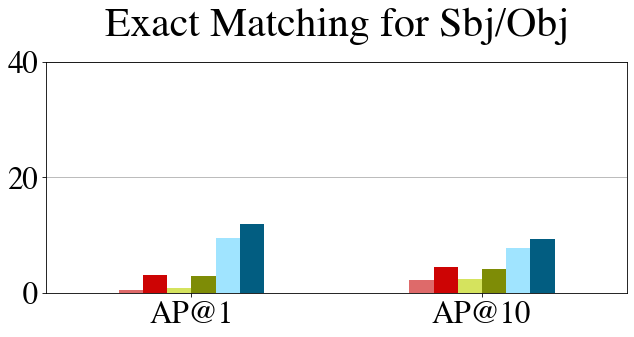}
%   \caption{}
  \label{fig:avg_prec3}
\end{subfigure}%
\begin{subfigure}{.19\textwidth}
%   \centering
  \includegraphics[width=\linewidth]{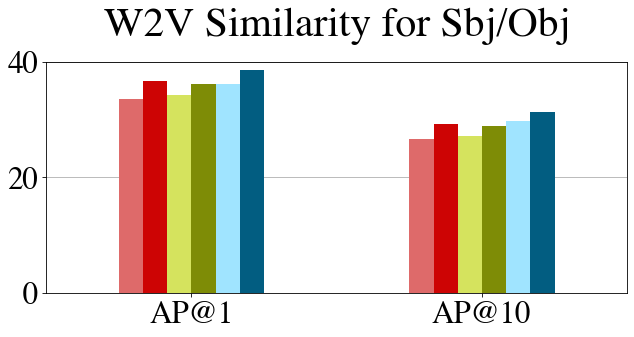}
%   \caption{}
  \label{fig:avg_prec4}
\end{subfigure}
\begin{subfigure}{.19\textwidth}
%   \centering
  \includegraphics[width=\linewidth]{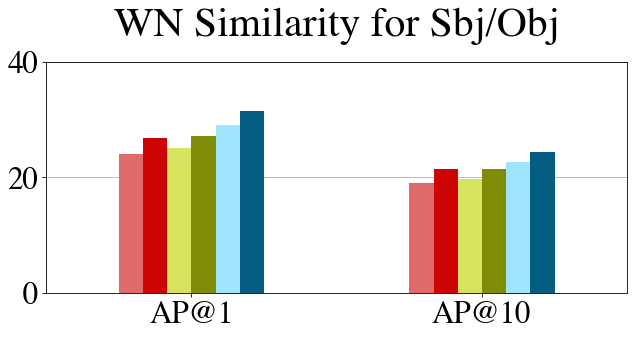}
%   \caption{}
  \label{fig:avg_prec5}
\end{subfigure}
\caption{\textbf{Soft Average Precision calculated on the tail classes on GQA-LT dataset using a variety of metrics.} We visualize results using exact similarity metrics, W2V-VG, and average of 6 WordNet metrics. The models using VilHub show consistently superior performance on the tail, when compared to similar models without the VilHub.}
\label{fig:avg_prec}
\end{figure*}

% triplet scores figure
\begin{figure*}
 \centering
\begin{subfigure}{\textwidth}
  \centering
  \includegraphics[width=.8\linewidth]{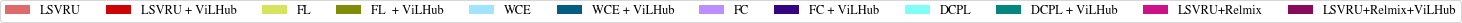}
\end{subfigure}
\begin{subfigure}{.24\textwidth}
%   \centering
  \includegraphics[width=\linewidth]{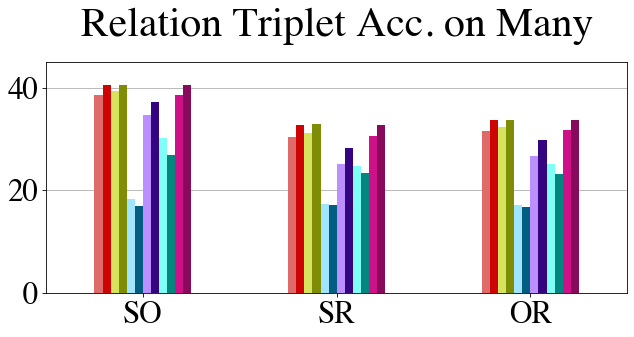}
%   \caption{}
  \label{fig:trip_so}
\end{subfigure}%
\begin{subfigure}{.24\textwidth}
%   \centering
  \includegraphics[width=\linewidth]{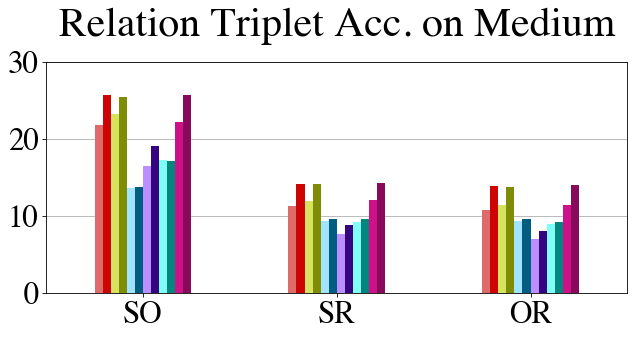}
%   \caption{}
  \label{fig:trip_sr}
\end{subfigure}
\begin{subfigure}{.24\textwidth}
%   \centering
  \includegraphics[width=\linewidth]{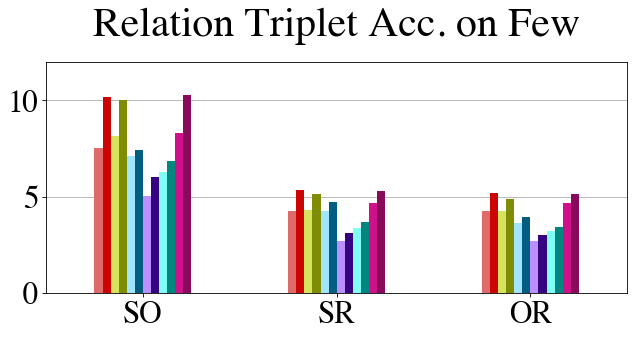}
%   \caption{}
  \label{fig:trip_or}
\end{subfigure}%
\vspace{-6mm}
\caption{Performance on relationship triplets (S, R, O) grouped by (S, O), (S, R), (O, R) on GQA-LT datasets}
\label{fig:triplets_pairs}
\end{figure*}
% hubness trends figure
\subsection{Compositional Results}
% \begin{figure}
% \centering
% % \begin{subfigure}{.15\textwidth}
%   \includegraphics[width=0.32\linewidth]{figures/comparisons/sbj_class_dist.png}
% % \end{subfigure}
% % \begin{subfigure}{.15\textwidth}
%   \includegraphics[width=0.32\linewidth]{figures/comparisons/rel_class_dist.png}
% % \end{subfigure}
% % \begin{subfigure}{.15\textwidth}
%   \includegraphics[width=0.32\linewidth]{figures/comparisons/triplet_dist.png}
% % \end{subfigure}
% \caption{Distribution of the number of examples for subjects/object, relations, and relationship triplet (SPO)}
% \label{fig:dist}
% \end{figure}
The long-tail poses a unique problem in relationship recognition due to the combinatorial nature of the problem.  \emph{dog}  and \emph{motorcycle} could be a common subject and object, but they may never be paired with the relationship \emph{ride}. %In standard classification and object detection there is only the distribution of the classes to worry about. %of the classes themselves along with the long-tail distribution of
In LTVRR, the long-tail distribution is not only on subjects, objects, and relations individually but also on their combinations. Meaning, we may not only have a rare combination of classes but also a combination of rare classes (e.g., $<$ otter, riding, dolphin $>$). %This poses a problem because even if a 2 classes are common, their combination might be rare, which makes recognizing a relationship between them difficult. 
%  This makes the whole triplet rare, even though the subject and object are common.
% \textcolor{red}{Fig~\ref{fig:dist} shows the stark difference in distribution between subjects/objects and relations and triplets.}
% \begin{figure}{\linewidth}
% \centering
% \includegraphics[width=0.7\linewidth]{figures/dist.png}
% \caption{\label{fig:dist}The distribution of classes of subjects/objects (upper), relations(middle), and triplets(bottom). All figures are plotted on a log-scale for y-axis}
% \end{figure}
Here we analyze how the performance is affected by this combinatorial nature of the problem.
Fig~\ref{fig:triplets_pairs} shows the performance on recognizing the entire triplet (S, R, and O) correctly on GQA-LT, which is the most important metric when evaluating relationship recognition. The results are grouped by pairs of (Subject, Object), (Subject, Relation), and (Object, Relation), and the accuracy is averaged over each group. VilHub improves existing approaches on this angle of performance; we can see in Fig.~\ref{fig:triplets_pairs} that the relationship triplet recognition performance is  exceeding 40\% when we group by pairs of (S, O), while it is under 35\% for (S, R) and (O, R). This shows how the more frequent (S,O) is more predictive of the entire triplet's performance (S,R,O) than other combinations.
Fig~\ref{fig:triplets_pairs} also shows the superior performance when using VilHub loss ($\approx$ 3\% gain ) for most of the models. This shows that VilHub also helps recognizing infrequent combinations of S, R, and O. Additionally, we can see that LSVRU+Relmix+VilHub gives the best performance on the \emph{few} category ($\approx$11\% on SO group) which shows the effectiveness of our augmentation strategy when combined with the VilHub regularizer.
% \input{tables/gqa_triplet_scores}

% \subsection{Ablation Studies}
% \noindent \textbf{RelMix on both subject and objects} here instead of either replacing the subject or the object, we replace both.
% \noindent \textbf{RelMix without ViLHub regularizer} here we use the RelMix strategy without the ViLHub regularization.

\label{sec:w2v_analyisis}
\label{sec:analysis} % ap analysis
\noindent \textbf{Soft Average Precision Analysis.} Fig~\ref{fig:avg_prec} shows the average precision of our models for the tail classes (least frequent 80\%). 
Concretely, we use the analysis to measure which models bring classes with similar meaning to the ground truth higher in the prediction rank.  In agreement with the human subject experiments, the results  reveal that all models are doing significantly better than the exact match metric suggests.
%This agrees with our observations when evaluating qualitative examples from all the models and suggests that using the exact ground truth matching is too harsh and to some extent can be suboptimal. 
Another takeaway from this analysis is that similarity metrics trained on relevant data are better at evaluating the models' performances than metrics trained on less relevant data. This can be seen when comparing the W2V-VG similarity metric (trained on VG) with the other metrics. W2V-VG metric is  $\approx$7\% more than wordNet metrics and $\approx$10\% more than W2V-GN (figures for W2V-GN in supp). Fig~\ref{fig:avg_prec} also shows a consistent improvement for models using VilHub in agreement with our previous results. Overall, these results imply that our models are better at bringing semantically relevant concepts higher in rank. We show similar observations for the head classes and analysis for RelMix in supp.

% We calculate the average precision of our models using soft metrics for the tail classes (least frequent 80\%). 
% Concretely, we use the analysis to measure which models bring classes with similar meaning to the ground truth higher in the prediction rank.  In agreement with the human subject experiments (provided in supp.), the results  reveal that all models are doing significantly better than the exact match metric suggests. We provide more details along with the full results of this analysis in the supp.
%This agrees with our observations when evaluating qualitative examples from all the models and suggests that using the exact ground truth matching is too harsh and to some extent can be suboptimal. 
% Fig~\ref{fig:avg_prec} also shows a consistent improvement for models using the ViLHub loss in agreement with our previous results. Overall, these results imply that our models are better at bringing semantically relevant concepts higher in the rank. We show similar observations for the head classes and analysis for Relmix in the supplementary.

% \section{Discussion}
% \label{sec:discussions}
% \input{LTVRR_ICCV21/LaTeX/discussions}

\section{Conclusion}
We proposed a new augmentation strategy, dubbed RelMix, and a visiolinguistic hubless regularizer (VilHub) to improve tail class performance in visual relationship recognition. We apply these approaches to the study of an important and challenging structured visual understanding problem, which aims to generalize visual relationship recognition task to the tail of the underlying distribution. We denote this problem as Long-Tail Visual Relationship Recognition (LTVRR), and we propose to study it on GQA-LT and VG8K-LT benchmarks that we built based on GQA and Visual Genome datasets. We implemented several SOTA baselines and applied them to this task. We showed that our novel adaptation of the VilHub regularizer and augmentation strategy (Relmix) improve the performance, especially for tail classes, while maintaining and sometimes improving performance on head classes. Additionally, our proposed methods are orthogonal to existing approaches and can be integrated with various SOTA models, improving their performance in most cases, as we have shown.

{\small
\bibliographystyle{ieee_fullname}
\bibliography{main}
}

\end{document}